\documentclass{svproc}
\usepackage{times}
\usepackage{upgreek}
\usepackage{nccmath}
\usepackage{mathtools}
\usepackage{array}
\usepackage{stfloats}
\input{macros}
\newcommand*{\scost}[1]{c_s(#1)\xspace} %
\newcommand*{\dcost}[1]{c_d(#1)\xspace} %
\newcommand*{\sv}[1]{s_{#1}} %
\newcommand*{\svk}[2][k]{s_{#2}^{#1}} %
\newcommand*{\ddk}[2][k]{d_{#2}^{#1}} %
\newcommand*{\zk}[2][k]{z_{#2}^{#1}} %

\usepackage{pgfplots}
\usepackage[caption=false]{subfig}
\pgfplotsset{compat=1.6}
\usepackage{color}
\usepackage[normalem]{ulem}  %

\begin{document}
\title{The Correlated Arc Orienteering Problem}

\author{Saurav Agarwal \and Srinivas Akella}
\authorrunning{S. Agarwal and S. Akella}
\institute{University of North Carolina at Charlotte, NC, USA 28223\\
\email{\{sagarw{10},sakella\}@uncc.edu}}
\maketitle

\begin{abstract}
	This paper introduces the correlated arc orienteering problem (CAOP), where the task is to find routes for a team of robots to maximize the collection of rewards associated with features in the environment. These features can be one-dimensional or points in the environment, and can have spatial correlation, i.e., visiting a feature in the environment may provide a portion of the reward associated with a correlated feature. A robot incurs costs as it traverses the environment, and the total cost for its route is limited by a resource constraint such as battery life or operation time. As environments are often large, we permit multiple depots where the robots must start and end their routes. The CAOP generalizes the correlated orienteering problem (COP), where the rewards are only associated with point features, and the arc orienteering problem (AOP), where the rewards are not spatially correlated. We formulate a mixed integer quadratic program (MIQP) that formalizes the problem and gives optimal solutions. However, the problem is NP-hard, and therefore we develop an efficient greedy constructive algorithm. We illustrate the problem with two different applications: informative path planning for methane gas leak detection and coverage of road networks.

	\keywords{Orienteering Problem, Informative Path Planning, Arc Routing}
\end{abstract}

\section{Introduction}
Consider a scenario in the aftermath of a natural disaster such as flooding.
A team of uncrewed aerial vehicles (UAVs) with cameras is deployed to assess the accessibility of a road network for emergency services.
The UAVs must traverse the line segments corresponding to the road network and use their cameras to capture images for analysis.
The {\em correlated arc orienteering problem} (CAOP), introduced in this paper, answers the following question:
How should routes for resource-constrained UAVs be planned such that the information gathered by the team along linear features is maximized by exploiting correlations between features?
The information related to the phenomenon, e.g., flooding, being monitored is modeled as {\em rewards}, which can be spatially correlated---flooding at a road segment may correlate with flooding of nearby low-lying road segments.
Furthermore, a UAV flying at a high enough altitude has a large camera field-of-view, and traversing a road segment may simultaneously capture images of nearby road segments.
The CAOP formulation can take advantage of these correlations to compute efficient routes for the robots.
Power lines and oil and gas pipelines have similar linear infrastructure, and such efficiencies can be exploited during inspections.

	\begin{figure}[t]
		\centering
		\hfill
		\subfloat[CAOP route]{%
		\includegraphics[height=0.25\textheight]{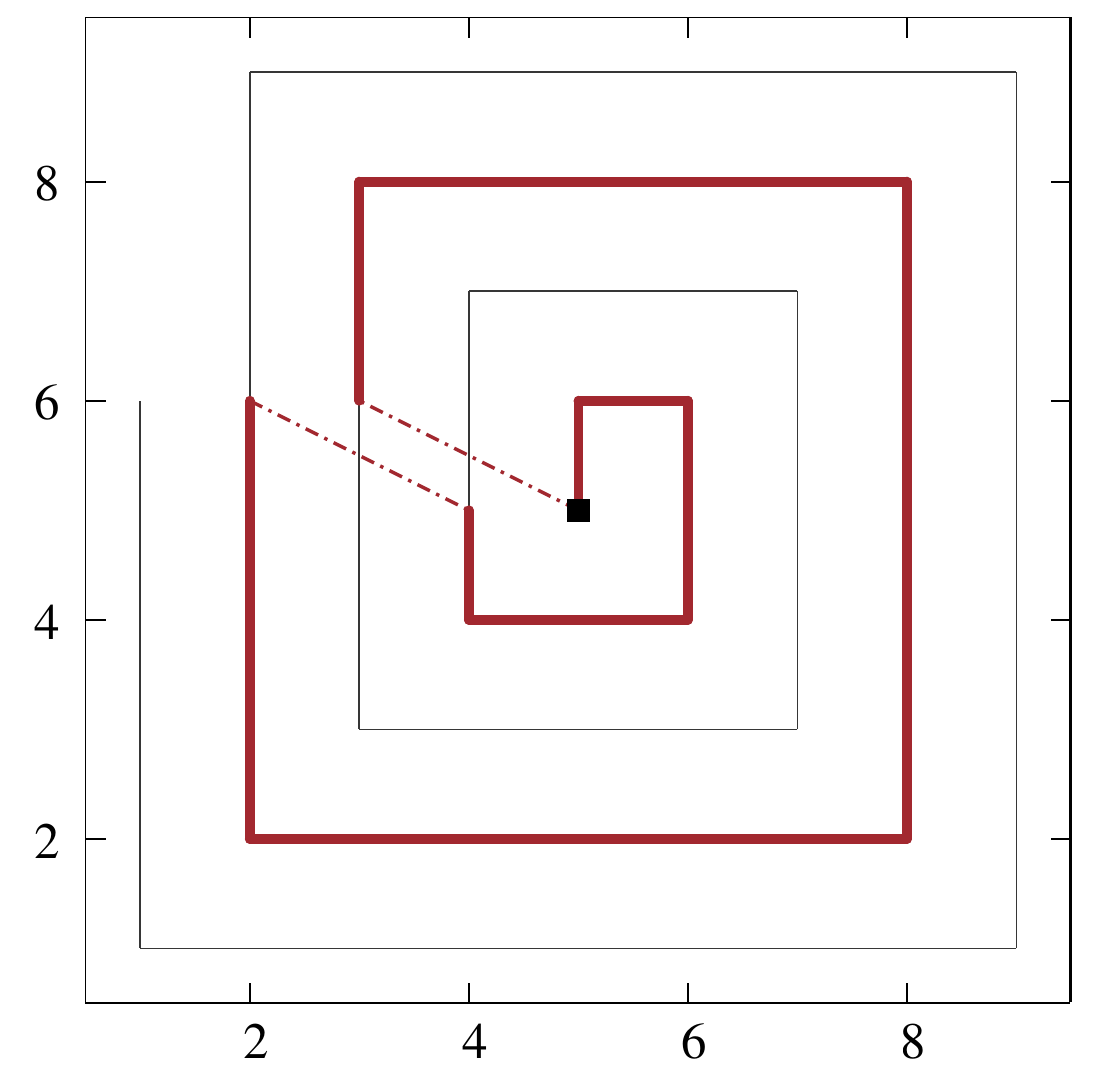}}
		\hfill
		\subfloat[AOP route]{%
		\includegraphics[height=0.25\textheight]{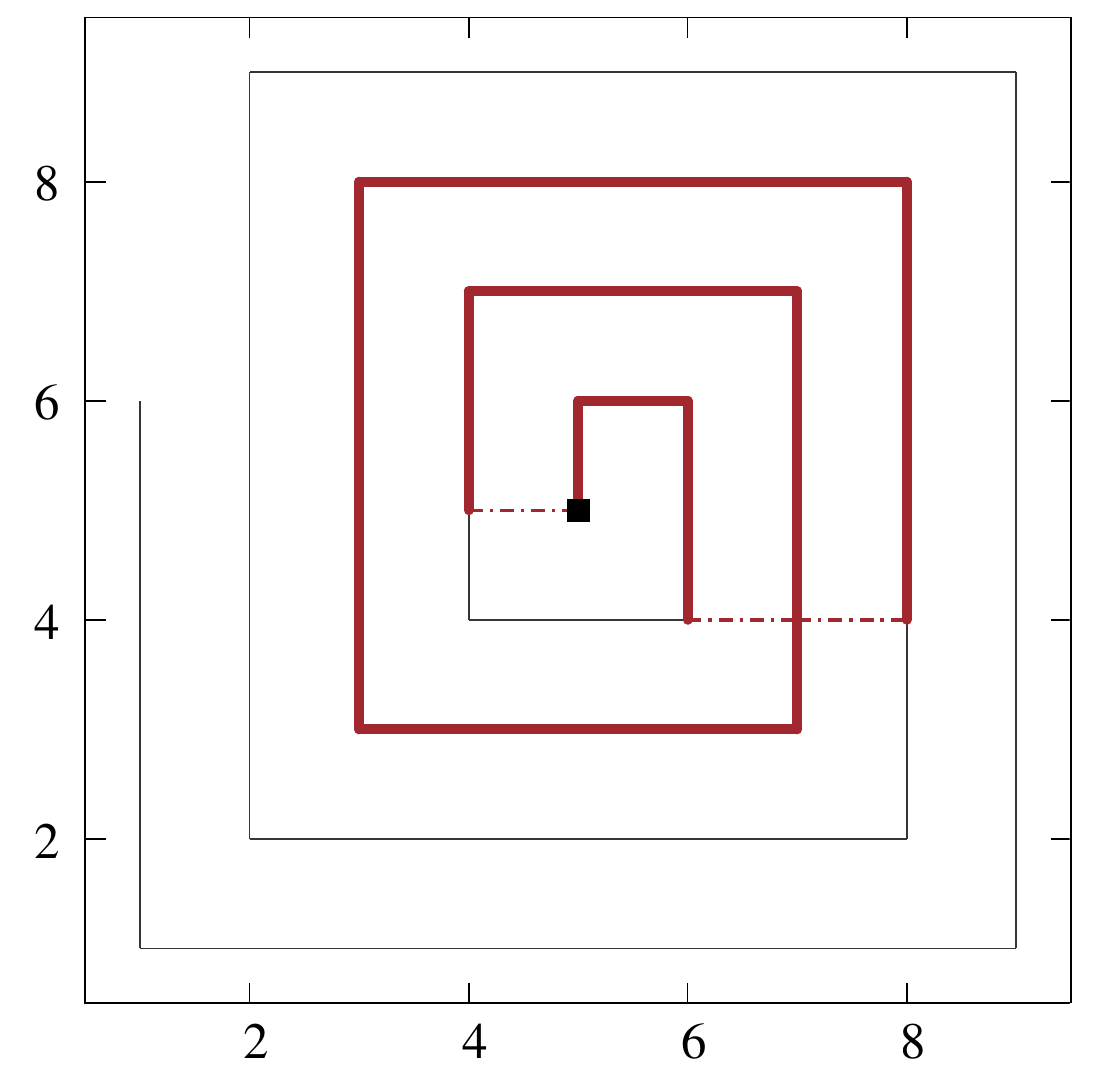}}
		\hfill
		\caption{Coverage of a spiral network, composed of 77 line segments of unit length, using a capacity-constrained robot.
			The rewards are the lengths of the edges, i.e., 1 for each constituent segment.
			The lateral field-of-view is set to 2 units---traversing an edge can fully cover the edges immediately parallel to it on both sides.
			The field-of-view models the correlation function.
			The black square represents the depot location.
			The solid red lines represent servicing of edges, while the dashed lines represent deadheading.
			Figure (a) shows the solution using the CAOP formulation.
			Figure (b) shows the solution for the AOP formulation, which does not use correlation information.
			With a resource capacity (budget) of 35 units, the AOP route covers 53 segments, while the CAOP covers 69 segments.
		\label{fig:example}}
	\end{figure}
The CAOP with multiple robots models the environment as a graph:
(1)~The linear features in the environment are represented by edges in the graph and have rewards associated with them.
(2)~The rewards are spatially correlated, i.e., traversing an edge provides a fraction of the reward from correlated edges.
(3)~The robots consume resources such as battery life or operation time while traversing the environment, and the total resource available to a robot is limited by a given {\em budget} or {\em capacity}.
The task is to find a set of routes for the robots that maximizes the total reward gathered while ensuring the total cost incurred by each robot is less than its capacity.
Since environments can be large and operating the robots from a single site may not yield reasonable solutions, we consider multiple sites where the robots can start and end their routes.
These sites are known as {\em depot} locations and are a subset of vertices in the underlying graph.
A unique characteristic of our formulation is that we consider two different modes of travel for the robots---servicing and deadheading.
A robot {\em services} an edge when it performs task-specific actions such as taking images along the edge.
A robot may traverse an edge without performing the servicing tasks, referred to as {\em deadheading}.
The robot may traverse faster during deadheading to optimize the operation time.
Having two modes of travel allows cost functions that depend on the travel mode, enabling algorithms to optimize the routes further.
\fgref{fig:example} shows an example of the CAOP for maximizing coverage of linear infrastructure with a capacity-constrained robot.
The lengths of the line segments model the rewards, while the sensor field-of-view models the correlations between line segments.

Our inspiration to study this problem comes from the correlated orienteering problem (COP) proposed by Yu et al.~\cite{YuSR16}.
However, the COP and related orienteering problems assume that the rewards are only available at the point features and, thus, are limited in their ability to model networks with linear features.
When the features of interest are one-dimensional {\em linear features}, they can be modeled as arcs or edges in a graph rather than vertices.
Such problems belong to the broader class of arc routing problems~\cite{CorberanL14}.
One such problem is the team orienteering arc routing problem (TOARP)~\cite{ArchettiSCSP14}, where the task is to maximize the reward by visiting arcs and edges in a graph using a team of vehicles.
Both point and linear environment features can be conveniently modeled in the CAOP formulation, and the CAOP considers the correlation between these features.
Thus, the CAOP generalizes both the COP and the TOARP. As the COP and the TOARP are NP-hard, the CAOP is also NP-hard.

This paper introduces the CAOP and formally states it as a mixed integer quadratic program (MIQP) that gives optimal solutions.
The MIQP formulation models rewards on the edges of a graph while incorporating spatial correlations between the edges.
As the problem is NP-hard, we develop a greedy constructive algorithm that generates solutions efficiently.
We illustrate the problem with two different applications: informative path planning for methane gas leak detection and coverage of road networks.

\section{Related Work}
\label{sc:related_work}
The correlated arc orienteering problem (CAOP) is related to orienteering problems, arc routing problems, and informative path planning (IPP).

\subsubsection{Orienteering and Related Problems:}
In the {\em orienteering problem} (OP) introduced by Golden et al.~\cite{GoldenLV87}, we are given a set of vertices with rewards (or scores) at each vertex and a time budget.
The task is to find a path from a start point to an endpoint through a subset of vertices to maximize the total reward collected while respecting the budget.
The OP is usually modeled as a graph, where the edges are the segments that can be used to form a path, and the edge weights model the time required to traverse the path.
A variant of the problem is to find a tour starting and ending at a given vertex.
The problem has elements of the knapsack problem and the traveling salesperson problem (TSP).
Chekuri et al.~\cite{ChekuriKP12} presented a $(2+\epsilon)$-approximation algorithm for the OP.
In the {\em team orienteering problem} (TOP), the task is to find $K$ paths (or tours), each limited by the time budget.
There are three common themes to the solution approaches.
Exact algorithms are often based on mixed integer linear programs (MILP) and use methods for branch-and-bound with cutting planes.
Metaheuristic approaches use techniques such as simulated annealing and memetic algorithms.
Since the problem is NP-hard, various heuristic and approximation algorithms have also been proposed.
These approaches and several other variants of the OP are covered in the survey by Gunawan et al.~\cite{GunawanLV16}.

The OP and its variants (e.g., the set orienteering problem~\cite{PenickaFS19}) assume that the rewards at the vertices are mutually independent.
Yu et al.~\cite{YuSR16} considered applications where the rewards may be spatially correlated, and introduced the {\em correlated orienteering problem} (COP).
In addition to the rewards at the vertices, a correlation function is given as input, which encapsulates the idea that visiting a vertex can provide information associated with a correlated vertex.
More concretely, a quadratic relationship is established through the correlation function, and the problem is termed the quadratic COP; we refer to this problem as the COP.
The advantages of using COP over OP were established, and an application to persistent monitoring was described.
A mixed integer quadratic program (MIQP) was developed to obtain optimal solutions.
The OP and the COP are vertex-based problems as the rewards are only associated with the vertices.
However, in some robotics applications, the rewards may be associated with linear features, modeled as edges in the graph.
The CAOP can handle rewards on both vertices and edges, thus generalizing the OP and the COP.
We provide an MIQP formulation and a greedy constructive algorithm, which also apply to the OP and the COP.

\subsubsection{Arc Routing Problems:}
Routing on arcs and edges in a graph belongs to the class of arc routing problems (ARPs)~\cite{CorberanL14}, widely studied in the operations research community.
The ARPs are used for planning vehicle routes for snow plowing, urban mail delivery, and salt spreading, where the linear features of the environment, such as road segments, are modeled as edges in a graph.
Such tasks can be automated using autonomous ground robots.
The {\em rural postman problem} (RPP) for a single robot and the {\em capacitated arc routing problem} (CARP) for multiple robots are two of the most widely used formulations for such applications.
{\em Line coverage}---coverage of linear infrastructure such as road networks and powerlines---was addressed by Agarwal and Akella~\cite{AgarwalA20ICRA}.
In these problems, which are generally NP-hard, a set of required edges that must be traversed is given.
The task is to find routes that minimize the total cost of travel while respecting the capacity constraint.
The {\em arc orienteering problem} (AOP)~\cite{GavalasKMPV16} differs from these coverage problems; the objective is to maximize the reward collected along the traversed edges of the route, similar to the orienteering problem but on arcs and edges instead of vertices.
The {\em team orienteering arc routing problem} (TOARP)~\cite{ArchettiSCSP14} is the multi-robot variant of the AOP.
Branch-and-bound with cutting planes and metaheuristic algorithms have been widely used for ARPs~\cite{MouraoP17}.
However, fast and efficient heuristic algorithms for robotics applications have not achieved a maturity comparable to vertex-based problems.
The CAOP generalizes the TOARP as it additionally handles correlations between the linear features.

\subsubsection{Informative Path Planning} (IPP) problems consider path planning for vehicles and robots to characterize and monitor the environment.
The information gathered using sensors is associated with the phenomenon to be observed and analyzed.
The task in IPP is to find robot paths that maximize an information metric while respecting capacity constraints.
Information gain, mutual information, and expected entropy reduction are commonly used information metrics.
Hollinger and Sukhatme~\cite{HollingerS14} proposed sampling-based techniques, combined with branch-and-bound, to generate a trajectory for a single robot for efficient information gathering with motion constraints.
Singh et al.~\cite{SinghKGK09} proposed a recursive greedy algorithm for IPP with a single robot and extended it for multiple robots.
The algorithm is exponential in the recursive depth and, therefore, can be computationally expensive if the instance size is large.
The IPP is closely related to orienteering problems.
Bottarelli et al.~\cite{BottarelliBBF19} use an algorithm for orienteering as a subroutine for IPP.
Like Binney et al.~\cite{BinneyKS13}, they discuss the advantage of continuous data sampling along edges.
These motivate formulating IPP as an arc routing-based CAOP.
Moreover, CAOP captures correlation, which is relevant for environmental monitoring when a phenomenon is spatially correlated.

\section{CAOP: Problem Statement}
\label{sc:problem_statement}
This section formally defines the {\em correlated arc orienteering problem} (CAOP).
The environment comprises linear features (line segments or curves) that contain information that needs to be collected by a team of $K$ robots.
We model the environment as an undirected and connected graph $G=(V, E)$, where $E$ is the set of edges representing linear features and $V$ is the set of vertices consisting of edge endpoints, edge intersections, and depot locations.
If we also have point features $V_f\subseteq V$ in the environment, we add an artificial edge $(v, v)$ for each point feature $v\in V_f$.
The {\em depots} $V_d\subseteq V$ are a subset of vertices at which the robots start and end their routes.

The task is to compute a set of routes $\Uppi = \{\pi^1,\ldots,\pi^K\}$ for $K$ robots.
A route for a robot $k$ must start and end at a specified depot $v^k_d \in V_d$.
There are two modes of travel for a robot---servicing and deadheading.
A robot is said to be {\em servicing} when task-specific actions such as taking images are performed along an edge.
We associate two binary variables, $\svk{ij}$ and $\svk{ji}$, with servicing an edge $(i, j) \in E$ by robot $k$: if a robot $k$ {\em services} edge $e$ in the direction $i\rightarrow j$, then $\svk{ij}$ is $1$ and $0$ otherwise; similarly, for the direction $j\rightarrow i$.
We define $\svk{e}= (\svk{ij} + \svk{ji})$ to denote whether an edge $e = (i, j) \in E$ is serviced by the robot $k$, and $\sv{e} = \sum_{k=1}^K\svk{e}$ for servicing by any of the robots.
An edge with a positive reward can be serviced at most once.
If a robot traverses an edge without servicing it, the robot is said to be {\em deadheading}.
We associate two non-negative integer variables, $\ddk{ij}$ and $\ddk{ji}$, with deadheading an edge $e = (i, j) \in E$ by robot $k$.
We define~$\ddk{e}$ similar to the service variables to denote total deadheadings of an edge by robot $k$.
An edge may be deadheaded as many times as required.

Each robot is constrained by a resource such as travel time, total travel distance, or battery life.
Such a constraint is represented by a {\em budget} or {\em capacity} $Q^k$ for a robot~$k$.
The consumption of resources is modeled by cost functions $\scost{e}$ for servicing and $\dcost{e}$ for deadheading an edge $e\in E$.
A robot may need to travel slower while performing task-specific actions such as taking images, resulting in a longer travel time for servicing than deadheading.
Thus, the two cost functions can differ.
The total cost incurred by a robot for a route $\pi^k$ must be less than its capacity $Q^k$:
\begin{align}
c(\pi^k) = \sum_{e\in E} \Big(\scost{e}\,\svk{e} + \dcost{e}\,\ddk{e}\Big) \leq Q^k\label{eqn:capacity}
.\end{align}

We follow the notation for correlated information in~\cite{YuSR16} and extend it to the CAOP.
A {\em reward} function $r: E\mapsto \mathbb R_{\geq 0}$ maps an edge to a non-negative measure of the information associated with the edge.
The spatial correlation between the edges is modeled by weight function $w: E \times E\mapsto \mathbb R_{\geq 0}$.
A weight of $w(e', e)$ is the fraction of the reward associated with an edge $e$ that can be obtained by traversing an edge $e',\;\forall e'\in E$.
It represents the {\em effectiveness} of obtaining information associated with the edge $e$ by traversing the edge $e'$.
The weight of an edge to itself is 0, i.e., $w(e, e) = 0,\;\forall e\in E$.
For each edge $e \in E$ with $r(e) > 0$, the set $N(e)$ is the set of edges with non-zero correlation with the edge $e$, i.e., $N(e) = \{e' \in E\mid w(e',e) > 0\}$.
In other words, the set $N(e)$ is the set of edges that can observe edge~$e$ through correlation.
Note that $w(e',e)$ need not be the same as $w(e, e')$, and $e'\in N(e)$ does not necessarily imply $e \in N(e')$.
For now, assume that the sum of the associated weights for an edge to be less than or equal to $1$, i.e., for each $e \in E$ with $r(e)> 0 $, $\sum_{e'\in N(e)} w(e',e) \leq 1$.
We will relax this constraint in the MIQP formulation and the constructive heuristic algorithm in Section~\ref{sc:algorithm}.
The total reward collected over all the routes in $\Uppi$ is given by:
\begin{align}
	\mc R(\Uppi) = \sum_{e\in E} \mathcal R(\Uppi, e) =\sum_{e\in E}r(e) \left(\sv{e} + \sum_{e'\in N(e)} w(e',e)\,\sv{e'}\left( \sv{e'} - \sv{e} \right)  \right) \label{eqn:path_utility}
.\end{align}
The quadratic term $\sv{e'}\left( \sv{e'} - \sv{e} \right)$ is $1$ if and only if $\sv{e'} = 1$ and $\sv{e} = 0$, for a pair of edges $e, e'$.
If the edge $e$ is serviced, i.e., $\sv{e} = 1$, the collected reward, associated with edge $e$, is $r(e)$, and the second term vanishes---there is no correlated contribution from neighboring edges.
If the edge $e$ is not serviced, the collected reward comes from servicing neighboring edges and is given by the second term.
The total reward $\mathcal R(\Uppi)$ collected over a set of routes $\Uppi$ for $K$ robots is to be maximized.

\begin{definition}
{\em Correlated arc orienteering problem (CAOP):}\\
Given a graph $G=(V, E)$, a reward function $r$, and a weight function $w$, the correlated arc orienteering problem with $K$ robots is to find a set of routes $\Uppi$ for the robots that maximizes the total reward collected $\mc R(\Uppi)$, while satisfying the capacity constraint for each robot.
\end{definition}

{\bf Relationship of COP and CAOP:}
In the above discussion, we modeled the linear features as edges under the presumption that the information is distributed only over the linear features.
In applications such as inspection of oil and gas pipelines, the oil wells may also be features of interest, i.e., point features.
Such point features can be conveniently incorporated into our formulation by creating an artificial edge for each point feature with a cost equivalent to that of inspecting the point feature.

The expression for total reward collected for a set of routes~\eqref{eqn:path_utility} is similar to the expression for COP in~\cite{YuSR16}.
One may then ask: Can we model the linear features as vertices in the graph and solve the COP?
While modeling total reward can be conveniently done in the COP for both types of features, transforming the routing constraints is challenging and leads to a substantial increase in the size of the instance~\cite{GavalasKMPV16}.
Furthermore, the graph structure is lost in the transformation, and the ability to incorporate practical aspects of a robotics application, e.g., asymmetric costs and kinematic constraints, becomes severely restricted.
Moreover, the original COP assumes that the sum of the associated weights for an edge is no more than $1$.
The MIQP formulation and the heuristic algorithm, as presented in Section~\ref{sc:algorithm}, place no such restriction on the weights.

\section{Exact and Heuristic Approaches for CAOP}
\label{sc:algorithm}
In this section, we present two approaches for solving the correlated orienteering problem (CAOP) with multiple robots.
We first develop a mixed integer quadratic program (MIQP) to obtain optimal solutions.
Thereafter, we develop a greedy constructive algorithm to solve the problem efficiently.

\subsection{Mixed Integer Quadratic Program for CAOP}
A mixed integer quadratic program (MIQP) is a formulation for optimization problems with integer and continuous variables, a quadratic objective function, and a set of linear constraints.
An MIQP provides a concise description of the CAOP, and solving the MIQP gives an optimal solution.
There are multiple solvers, e.g., Gurobi and CPLEX, that can solve MIQPs efficiently for small to moderate instance sizes.

\subsubsection{Variables:}
We have the following variables for the MIQP.
\begin{itemize}
	\item Binary service variables $\svk{ij}, \svk{ji} \in \{0, 1\}$ for each edge $(i, j)$ and each robot $k$.
	\item Integer deadheading variables $\ddk{ij}, \ddk{ji} \in \mathbb N\cup \{0\}$ for each edge $(i, j)$ and each robot~$k$.
	\item Integer flow variables $\zk{ij}, \zk{ji} \in \mathbb N\cup \{0\}$ for each edge $(i, j)$ and each robot $k$.
		The flow variables are used in connectivity constraints to ensure that routes are connected to their respective depots.
	\item Real variables $\omega_e\in \mathbb R_{\geq 0}$ for each edge $e\in E$.
\end{itemize}

\subsubsection{Objective Function:}
The objective is to maximize the total reward  collected over the entire set of routes as given in the expression for $\mathcal R(\Uppi)$ in~\eqref{eqn:path_utility}.
The number of non-linear terms in the expression is $\mathcal O(m^2)$, where $m$ is the number of edges.
The non-linear terms make it challenging to solve an MIQP, and therefore, we reduce the number of non-linear terms to $\mc O(m)$ by adding $m$ variables and constraints.
These $\omega_e$ variables can be interpreted as the cumulative weights corresponding to the rewards obtained by servicing the neighboring edges.
The total reward is now expressed as:
\begin{equation}
\begin{split}
	\mathcal R(\Uppi) =
	&\sum_{e\in E} \left[r(e) \Big(\sv{e} + \omega_e\left(1-\sv{e}\right) \Big) \right]\\
	\text{subject to }\quad &\omega_e \leq \sum_{e'\in N(e)} w(e',e)\,\sv{e'}\leq 1, \quad \omega_e \in \mathbb R_{\geq 0}, \; \forall e\in E
	\label{eqn:PiReward}
.\end{split}
\end{equation}
This formulation also allows us to remove the assumption that the sum of the weights is less than $1$, i.e., we no longer require $\sum_{e'\in N(e)} w(e',e) \leq 1$.
Note that $\omega_e$ will always take the value given by the summation if the sum is less than $1$, as the expression is part of a maximization objective function.

We modify this expression by scaling the reward by a factor $\lambda$ and subtracting the total cost of the routes.
Incorporating the total cost allows the algorithm to break ties between routes with the same reward.
The scaling factor $\lambda$ is large enough to add an edge if it provides a positive reward and the corresponding route is within capacity limits.
This criterion is satisfied by the ratio of the sum of the capacities and the minimum positive reward.
The scaling factor also ensures that the optimal set of routes have the maximum total reward.
From equations~\eqref{eqn:capacity} and~\eqref{eqn:PiReward}, the objective function can be written as:%
\begin{align*}
	\text{Maximize:}\quad\lambda\, \mathcal R(\Uppi) - \sum_{k=1}^K c(\pi^k), \quad \text{where } \lambda = \frac{\max \{Q^k,\;\forall k\}}{\min \{r(e) \mid e\in E, r(e) > 0\} } \quad
.\end{align*}

\subsubsection{Routing Constraints:}
The final piece of the MIQP formulation is the set of routing constraints that ensures connectivity of a route for each robot to the corresponding depot and the elimination of sub-tours.
The routing constraints can be expressed as a set of generalized flow constraints and symmetry constraints~\cite{AgarwalA20ICRA}.
For ease of notation, we define the following sets:%
\begin{gather*}
	\mathcal A=\;\smashoperator{\bigcup_{(i, j)\in E}} \;\{(i, j), (j, i)\},\quad
	H(\mathcal A, v) = \{(i, v)\in \mathcal A\}, \, \text{and}\quad
	T(\mathcal A, v) = \{(v, j)\in \mathcal A\}
.\end{gather*}
Here, $\mc A$ is the set of all arcs, and $H(\mathcal A, v)$ is the set of arcs in $\mathcal A$ that have $v\in V$ as the head.
Similarly, $T(\mathcal A, v)$ is the set of arcs that have $v$ as the tail.

We have the following set of {\em routing constraints} for each robot $k \in \{1,\ldots,K\}$:
\begin{align}
	&\text{flow from the depot: } &&\sumSr{(i, j)\in T(\mc A, v_d^k)} \zk{ij} \quad=\quad\sumS{(i, j)\in \mc A}\svk{ij}\label{eqn:flowDepot}\\
	&\text{flow conservation: } &&\sumS{(i, j)\in H(\mc A, v)}\zk{ij} \;- \sumSr{(i, j)\in T(\mc A, v)} \zk{ij} \;=\sumSr{(i, j)\in H(\mc A, v)}\svk{ij}, \quad \forall v \in V\setminus \{v_d^k\}\label{eqn:edgeFlow}\\
	& \text{limits on the flow: }&& \zk{ij}\quad\leq\quad\sumS{(i, j)\in \mc A}\svk{ij}, \quad \forall (i, j)\in \mc A\label{eqn:flowLimit2}\\
	& && \zk{ij}\quad\leq\quad\lvert E \rvert (\svk{ij} +\ddk{ij}), \quad \forall (i, j)\in \mc A\label{eqn:flowLimit1}\\
	&\text{vertex symmetry: } &&\sumSl{(i, j) \in H(\mc A, v)}(\svk{ij} + \ddk{ij}) \; = \sum_{(i, j) \in T(\mc A, v)}(\svk{ij} + \ddk{ij}) ,\quad\forall v \in V\label{eqn:symmetry}
\end{align}
The constraints~\eqref{eqn:flowDepot}--\eqref{eqn:flowLimit1} are generalized flow constraints that together ensure the connectivity of the route to the depot and prohibit any sub-tours.
The integer variables $\zk{ij}$ are flow variables for each edge direction.
Constraint~\eqref{eqn:flowDepot} defines the amount of flow released from the depot vertex $v_d^k$, which acts as a flow source.
For any vertex $v$ (other than the depot vertex), a flow equal to the number of servicing arcs, with $v$ as the head, is absorbed by the vertex.
This is expressed in constraints~\eqref{eqn:edgeFlow}.
The amount of flow through an arc is limited by constraints~\eqref{eqn:flowLimit2} and~\eqref{eqn:flowLimit1}.
Finally, the vertex symmetry constraints~\eqref{eqn:symmetry} ensure that the number of arcs entering a vertex is the same as the number of arcs leaving it.

\subsubsection{MIQP Formulation for CAOP:}
We can now pose the CAOP as an MIQP:%
\begin{equation}
\text{Maximize:}\quad	\lambda \sum_{e\in E} \left[r(e) \Big(\sv{e} + \omega_e\left(1-\sv{e}\right) \Big) \right]
	- \sum_{k=1}^K c(\pi^k)
\end{equation}
\hspace{12ex}subject to:%
\begin{align}
	&\sv{e} = \sum_{k=1}^{K} \svk{e} = \sum_{k=1}^{K} (\svk{ij} + \svk{ji}) \leq 1,\quad \forall e=(i, j)\in E\\
	&\omega_e \leq \sum_{e'\in N(e)} w(e', e)\,\sv{e'},\quad \forall e\in E
\end{align}
\begin{align}
	&c(\pi^k) = \sum_{e\in E} \Big(\scost{e}\,\svk{e} + \dcost{e}\,\ddk{e}\Big) \leq Q^k, \quad \forall k\in \{1,\ldots,K\} \\
	&\text{routing constraints \eqref{eqn:flowDepot}--\eqref{eqn:symmetry} for each robot } k \in \{1,\ldots,K\}\\
	&0\leq \omega_e\leq 1,\quad \omega_e\in \mathbb R,\; \forall e\in E\\
	&\svk{ij}, \svk{ji}\in \{0, 1\}, \quad \forall (i, j) \in E\\
	&\ddk{ij}, \ddk{ji}, \zk{ij}, \zk{ji} \in \mathbb N \cup \{0\}, \quad \forall (i, j) \in E
\end{align}

{\em Anytime property of the MIQP:}
Similar to COP, the trivial solution with empty routes is a feasible solution to the MIQP.
Commercial solvers for MIQP maintain the best feasible solution and a lower bound to improve the quality of the solution.
Such a solver has the anytime property---a feasible solution can be obtained by interrupting the execution at any time.
However, a substantial computational effort is spent in obtaining a good lower bound by solving linear relaxations and applying cutting planes that improve the quality of the polyhedron of the relaxation.
It can take a long time before a meaningful feasible solution is obtained, especially for large graphs with several robots.
Thus, the anytime property of the MIQP is often not valuable for robotics applications that require rapid solutions.
This motivates us to develop a fast heuristic algorithm.
We also use the solutions obtained from the heuristic algorithm to provide a good initial solution to the MIQP solver.
Nevertheless, MIQP serves the essential purpose of defining the problem concretely, obtaining optimal solutions for small instances, and benchmarking heuristic algorithms by comparing the quality of the solutions.

\subsection{A Greedy Constructive Algorithm}
The correlated arc orienteering problem (CAOP) is NP-hard in general, and solving the MIQP to obtain optimal solutions can take a long time.
In this section, we develop a greedy constructive algorithm, given in Algorithm~\ref{alg:caop}.
The algorithm has three main steps performed iteratively:
maintain a set of routes for the robots, greedily select an edge to be added to a route, and efficiently construct a new route by adding the selected edge.
The input to the algorithm is a graph $G=(V, E)$, the set of depots $V_d\subseteq V$, the number of robots~$K$, the reward function~$r$, and the weight function~$w$.
The output of the algorithm is a set of routes $\Uppi=\{\pi^1,\ldots,\pi^K\}$.
\subsubsection{Initialization:}
The routes are represented as a sequence of service arcs and are empty initially.
A set $\mc P^k$ of potential edges that can be added to a route $k$ is maintained.
For each edge $e\in E$, two sets of edges---neighbors $N(e)$ and co-neighbors $\overline N(e)$---are maintained (lines 3--4).
An edge $e'$ is a {\em neighbor} of an edge $e$ if servicing $e'$ can provide some reward associated with  $e$, i.e., $e'\in N(e)$ if $w(e', e)>0$, as discussed in Section~\ref{sc:problem_statement}.
An edge $\bar e$ is a {\em co-neighbor} of $e$ if servicing $e$ provides some reward associated with $\bar e$.
We define {\em utility} $\mathcal U$ as the net reward that can be obtained by servicing an edge due to its own reward and correlated rewards from its co-neighbors (line 5).  
Next, we iterate over the edges and compute the route for servicing an edge, i.e., the cost of going from the depot vertex $v_d^k$ to the tail $t_e$ of the edge, servicing the edge, and coming back to the depot from the head $h_e$ of the edge (line 8).
If the cost of the route is less than the capacity of the robot $k$, we add the edge to the list of potential edges~$\mc P^k$ (line 9).

\subsubsection{Greedy Construction:}
To dynamically compute the scaling factor $\lambda$, we maintain the minimum utility $u_{\min}$ and the maximum incremental cost $c_{\max}$ of adding edges to the routes (lines 12--13).
The core part of the algorithm is to iterate over the potential edges~$\mc P^k$ for each route~$k$ and find the edge that gives the best value based on a {\em greedy criterion} (lines~15--21).
The criterion for selecting the optimal edge to add to a route is the difference between the scaled utility and the cost of adding the edge (line~20).
The scaling factor ensures that if an edge has a positive utility, the value $u$ is non-negative, and the edge can be added to the route.
We dynamically update the value of $\lambda$ instead of setting it to a large value to ensure the scaled utility does not dominate the cost of adding an edge.
If no edge with a non-negative value is found, the algorithm terminates (line~22); otherwise, the edge with the largest value based on the greedy criterion is selected.
The service variable~$s_e$ for the selected edge is set to $1$, and the corresponding route is updated (line~23).
As we have serviced the selected edge, the neighboring edges $N(e)$ cannot get any reward associated with $e$ through correlation.
Hence, the utility of the neighboring edges is reduced by the reward they could obtain if the selected edge were not serviced (lines~24--25).
Similarly, servicing the selected edge gave us rewards through the correlation of co-neighbors.
If it so happens that an edge from the co-neighbors is selected in the future, the correlated reward would be received twice.
Therefore, the utility of the co-neighbors is appropriately reduced (lines~26--27).
Additionally, the neighbors of the co-neighbors may require an update depending on the correlation function (line~28).
Finally, the updated route needs to modify the set of potential edges and compute new potential routes (lines~30--32).
Any edge that cannot be added to the route under the capacity constraint is removed from the list of potential edges.

\subsubsection{Complexity Analysis:}
Let $m$ be the number of edges in the graph.
The initialization of the neighbors, co-neighbors, and utilities take $\mc O(m^2)$ computation time (lines~2--5).
In lines~6--9, the computation of routes for a single edge takes constant time, and the complexity of initializing potential edges is $\mc O(Km)$.

Using a linked-list for the potential edges $\mc P^k$, neighbors $N(e)$, and co-neighbors $\overline N(e)$, the removal of an element can be done is constant time while iterating through the list (line~18).
In each iteration, selecting an edge with the optimal value based on the greedy criterion can be done in $\mc O(Km)$ computation time (lines~15--21).
Similarly, updating neighbors and co-neighbors can be done in linear time $\mc O(m)$ (lines~24--27), and in time $\mc O(m^2)$ if line~28 is required.
Finally, updating potential edges involves calling \textsc{ComputeRoute} at most $m$ times in each iteration.
The complexity of this step thus depends on the complexity of the algorithm for computing routes.
As we will discuss in the next section, we develop an algorithm that can construct a new route by adding an edge to an existing route in $\mc O(l)$ computation time, where $l$ is the number of edges (or arcs) in the existing route.
Using such an algorithm for computing routes, the complexity of updating potential edges is $\mc O(m^2)$.
If the capacity is large enough, all $m$ edges could be added to the set of routes.
Thus, the main while loop (line~10) may be executed $m$ times, and the overall complexity of the algorithm is $O(m^3)$.

\begin{algorithm}[htbp]
	\LinesNumbered
	\KwData{Graph $G=(V, E)$, depot $V_d\subseteq V$, $K$, reward $r(\cdot)$, weight $w(\cdot\,,\cdot)$}
	\KwResult{Routes $\Uppi=\{\pi^1,\ldots,\pi^K\} $}
	$\pi^k\gets\emptyset,\, \mc P^k\gets\emptyset, \quad k=\{1,\ldots,K\}$\;
	\For{$e\in E$} {
		$N(e) \gets \{e'\in E \mid w(e', e) > 0\}$\tcp*{Neighbors of $e$}
		$\overline N(e) \gets \{e'\in E \mid w(e, e') > 0\}$\tcp*{Co-neighbors of $e$}
		$\mc U(e)\gets r(e) + \sum_{e'\in \overline N(e)} w(e, e')\; r(e')$\tcp*{Utility of $e$}
	}
	\For{$e\in E$} {
		\For{$k=1$ \KwTo $K$}{
			$c(\rho_e^k) \gets c(v_d^k, t_e) + c(e) + c(h_e, v_d^k)$\tcp*{Route with edge $e$}
			\lIf(\tcp*[f]{Potential edges $\mc P^k$}){$c(\rho_e^k)\leq Q^k$}{$\mc P^k.\textsc{Push}(e)$}
		}
	}
	\While{{\em \textsc{true}}}{
		$u^* \gets -\infty$;\\
		$u_{\min} \gets \min\,\{\mc U(e) \mid e\in E,\;s_e\neq 1\}$\tcp*{Min utility}
		$c_{\max} \gets \max\{c(\rho_e^k) - c(\pi^k) \mid e\in \mc P^k,\;s_e\neq 1, \forall k\}$\tcp*{Max cost}
		$\lambda \gets {c_{\max}}/{u_{\min}}$\tcp*{Scaling factor}
		\For{$k=1$ \KwTo $K$} {
			\For(\tcp*[f]{Iterate over potential edges}){$e\in \mc P^k$} {
				\If{$\sv{e}= 1$}{Remove $e$ from $\mc P^k$;}
				\Else{
					$u \gets \lambda\, \mc U(e) - \left(c(\rho_e^k) - c(\pi^k)\right)$\tcp*{Greedy criterion}
					\lIf{$u > u^*$} {$u^* \gets u;\; k^* \gets k;\; e^*\gets e$}
				}
			}
		}
		\lIf{$u^* < 0$}{\Break}
		$\sv{e^*}\gets 1;\; \pi^{k^*}\gets \rho_{e^*}^{k^*}$\tcp*{Update route $\pi^{k^*}$}
		\For{$e'\in N(e^*)$} {
			$\mc U(e') \gets \max \{0,\;\mc U(e') - w(e', e^*)\;r(e^*)\} $\tcp*{Update neighbors}
		}
		\For{$\bar e\in \overline N(e^*)$} {
			$\mc U(\bar e) \gets \max \{0,\;\mc U(\bar e)  - w(e^*, \bar e)\;r(\bar e)\} $\tcp*{Update co-neighbors}
			Update neighbors of $\bar e$, if required\;
		}
		Remove $e^*$ from the list of neighbors and co-neighbors for all edges\;
		\For{$e\in \mc P^{k^*}$} {
			$\rho_e^{k^*} \gets \textsc{ComputeRoute}(G, \pi^{k^*}, v_d^{k^*}, e)$\tcp*{Compute new routes}
			\lIf{$c(\rho_e^{k^*}) > Q^{k^*}$}{Remove $e$ from $\mc P^{k^*}$}
		}
	}
	\caption{A greedy constructive algorithm for CAOP}
	\label{alg:caop}
\end{algorithm}

\subsection{\textsc{ComputeRoute}: Constructive Edge-Insertion Routing Heuristic}
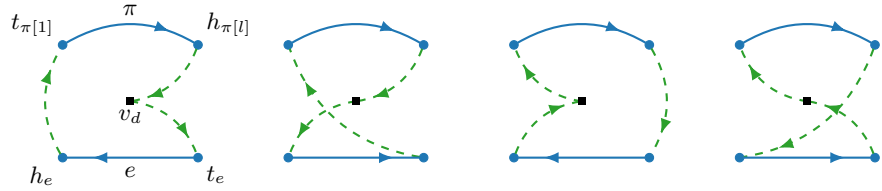
\begin{figure}[htbp]
	\centering
	\begin{center}
	\begin{tikzpicture}[scale=0.5]
		\usetikzlibrary{calc}
		\tikzstyle{scEnd}=[near end, fill=none, mDarkRed]
		\tikzstyle{scSt}=[near start, fill=none, black]
		\tikzstyle{scMid}=[midway, fill=none, black]

		\coordinate (0) at (0,0);
		\coordinate (da) at (-0.1,-0.1);
		\coordinate (db) at (0.1,0.1);

		\node[place] (1) at (-1.8,  1.5)		{};
		\node[place] (2) at ( 1.8,  1.5)			{};
		\node[place] (3) at (-1.8, -1.5)		{};
		\node[place] (4) at ( 1.8, -1.5)		{};

		\node[fill=none, below]	(d)			at (0,0)	{$v_d$};
		\node[fill=none, above left]		at (1)		{$t_{\pi[1]}$};
		\node[fill=none, above right]		at (2)		{$h_{\pi[l]}$};
		\node[fill=none, below left]		at (3)		{$h_{e}$};
		\node[fill=none, below right]		at (4)		{$t_{e}$};

		\path[mBlue] (1)  edge [mid2arc, bend left]  node[scMid, above]  {$\pi$} (2);
		\draw[mid2arc, mBlue] (4) -- (3)  node [scMid, below] {$e$};
		\draw[mid2arc, mGreen, dashed] (0) to [bend left] (4);
		\draw[mid2arc, mGreen, dashed] (3) to [bend left] (1);
		\draw[mid2arc, mGreen, dashed] (2) to [bend left] (0);

		\fill[black] ($ (0) + (da) $) rectangle ($ (0) + (db) $);

		\coordinate (off) at (6, 0);
		\foreach \x in {1,...,4} {
			\node[place] (\x) at ( $ (\x) +  (off) $)			{};
		}
		\coordinate (0) at ($ (0) + (off) $);
		\draw[mid2arc, mGreen, dashed] (off) to [bend right] (3);
		\draw[mid2arc, mGreen, dashed] (4) to [bend left] (1);
		\draw[mid2arc, mGreen, dashed] (2) to [bend left] (off);
		\fill[black] ($ (off) + (da) $) rectangle ($ (off) + (db) $);
		\path[mBlue] (1)  edge [mid2arc, bend left]  node[scMid, above]  {} (2);
		\draw[mid2arc, mBlue] (3) -- (4)  node [scMid, below] {};
	
		\foreach \x in {1,...,4} {
			\node[place] (\x) at ( $ (\x) +  (off) $)			{};
		}

		\coordinate (0) at ($ (0) + (off) $);
		\draw[mid2arc, mGreen, dashed] (0) to [bend left] (1);
		\draw[mid2arc, mGreen, dashed] (2) to [bend left] (4);
		\draw[mid2arc, mGreen, dashed] (3) to [bend left] (0);
		\fill[black] ($ (0) + (da) $) rectangle ($ (0) + (db) $);
		\path[mBlue] (1)  edge [mid2arc, bend left]  node[scMid, above]  {} (2);
		\draw[mid2arc, mBlue] (4) -- (3)  node [scMid, below] {};

		\foreach \x in {1,...,4} {
			\node[place] (\x) at ( $ (\x) +  (off) $)			{};
		}
	
		\coordinate (0) at ($ (0) + (off) $);
		\draw[mid2arc, mGreen, dashed] (0) to [bend left] (1);
		\draw[mid2arc, mGreen, dashed] (2) to [bend left] (3);
		\draw[mid2arc, mGreen, dashed] (4) to [bend right] (0);
		\fill[black] ($ (0) + (da) $) rectangle ($ (0) + (db) $);
		\path[mBlue] (1)  edge [mid2arc, bend left]  node[scMid, above]  {} (2);
		\draw[mid2arc, mBlue] (3) -- (4)  node [scMid, below] {};

	\end{tikzpicture}
\end{center}
	\caption{Four different ways of adding an edge $e$ at either end of an existing route $\pi$.}
	\label{fig:ends}
\end{figure}

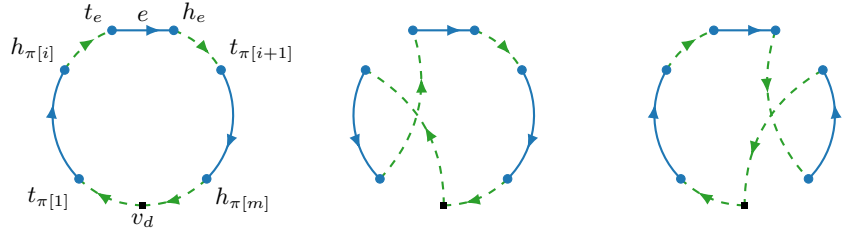
\begin{figure}[htbp]
	\centering
	\begin{center}
	\begin{tikzpicture}[scale=0.4]
		\usetikzlibrary{calc}
		\tikzstyle{scEnd}=[near end, fill=none, mDarkRed]
		\tikzstyle{scSt}=[near start, fill=none, black]
		\tikzstyle{scMid}=[midway, fill=none, black]

		\coordinate (0) at (0,0);
		\coordinate (da) at (-0.1,-0.1);
		\coordinate (db) at (0.1,0.1);
		\def\radi{3}

		\coordinate (dd) at (-90:\radi);
		\coordinate (1) at (225:\radi)		{};
		\coordinate (2) at (150:\radi)		{};
		\coordinate (3) at (110:\radi)		{};
		\coordinate (4) at (70:\radi)		{};
		\coordinate (5) at (30:\radi)		{};
		\coordinate (6) at (-45:\radi)		{};

		\node[fill=none, below]	(d)			at (-90:\radi)	{$v_d$};
		\node[fill=none, below left]		at (1)		{$t_{\pi[1]}$};
		\node[fill=none, above left]		at (2)		{$h_{\pi[i]}$};
		\node[fill=none, above left]		at (3)		{$t_{e}$};
		\node[fill=none, above right]		at (4)		{$h_{e}$};
		\node[fill=none, above right]		at (5)		{$t_{\pi[i+1]}$};
		\node[fill=none, below right]		at (6)		{$h_{\pi[m]}$};

		\draw[mid2arc=0.7, mBlue] (1) arc (225:150:\radi);
		\draw[mid2arc, mBlue] (3) -- (4)  node [scMid, above] {$e$};
		\draw[mid2arc=0.7, mBlue] (5) arc (30:-45:\radi);
		\draw[mid2arc=0.7, mGreen, dashed] (dd) arc (270:225:\radi);
		\draw[mid2arc=0.7, mGreen, dashed] (2) arc (150:110:\radi);
		\draw[mid2arc=0.7, mGreen, dashed] (4) arc (70:30:\radi);
		\draw[mid2arc=0.7, mGreen, dashed] (6) arc (-45:-90:\radi);

		\fill[black] ($ (dd) + (da) $) rectangle ($ (dd) + (db) $);

		\foreach \x in {1,...,6} {
			\node[place] (\x) at (\x)			{};
		}

		\coordinate (off) at (10, 0);
		\foreach \x in {1,...,6} {
			\coordinate (\x) at ( $ (\x) +  (off) $)			{};
		}

		\coordinate (d) at ($ (d) + (off) $);
		\coordinate (dd) at ($ (dd) + (off) $);

		\draw[mid2arc=0.7, mBlue] (2) arc (150:225:\radi);
		\draw[mid2arc, mBlue] (3) -- (4)  node [scMid, below] {};
		\draw[mid2arc=0.7, mBlue] (5) arc (30:-45:\radi);
		\draw[mid2arc=0.5, mGreen, dashed] (dd) to [bend right=30] (2);
		\draw[mid2arc=0.7, mGreen, dashed] (1) to [bend right=30] (3);
		\draw[mid2arc=0.7, mGreen, dashed] (4) arc (70:30:\radi);
		\draw[mid2arc=0.7, mGreen, dashed] (6) arc (-45:-90:\radi);

		\fill[black] ($ (dd) + (da) $) rectangle ($ (dd) + (db) $);

		\foreach \x in {1,...,6} {
			\node[place] (\x) at (\x)			{};
		}
		\coordinate (off) at (10, 0);
		\foreach \x in {1,...,6} {
			\coordinate (\x) at ( $ (\x) +  (off) $)			{};
		}

		\coordinate (d) at ($ (d) + (off) $);
		\coordinate (dd) at ($ (dd) + (off) $);

		\draw[mid2arc=0.7, mBlue] (1) arc (225:150:\radi);
		\draw[mid2arc, mBlue] (3) -- (4)  node [scMid, below] {};
		\draw[mid2arc=0.7, mBlue] (6) arc (-45:30:\radi);
		\draw[mid2arc=0.7, mGreen, dashed] (dd) arc (270:225:\radi);
		\draw[mid2arc=0.7, mGreen, dashed] (2) arc (150:110:\radi);
		\draw[mid2arc=0.4, mGreen, dashed] (4) to [bend right=30] (6);
		\draw[mid2arc=0.7, mGreen, dashed] (5) to [bend right=30] (dd);

		\fill[black] ($ (dd) + (da) $) rectangle ($ (dd) + (db) $);

		\foreach \x in {1,...,6} {
			\node[place] (\x) at (\x)			{};
		}
	\end{tikzpicture}
\end{center}
	\caption{Three of the eight different ways of adding an edge $e$ in the interior of an existing route $\pi$.}
	\label{fig:eightways}
\end{figure}

An efficient routing algorithm is essential for generating routes for the CAOP greedy constructive algorithm.
Given an existing route and an edge, the constructive edge-insertion heuristic inserts the edge in the route in time that is linear in the size of the route.
We represent a route $\pi$ by a sequence of $l$ service edges, and $t_e$ and $h_e$ represent the tail and the head for an edge $e$.
We observe that there are four ways of inserting an edge at either end of an existing route, as shown in \fgref{fig:ends}.
When the existing route has only one service edge and a new edge is added, there are four ways of forming a route and the algorithm will give the optimal solution.
When the existing route has two or more service edges and a new edge is added, there are eight possible ways of forming a route with the new edge inserted in the interior of the route.
Three of these eight ways are illustrated in \fgref{fig:eightways}.
In essence, the algorithm iterates over the edges of the route and splits the route into two halves, around the depot.
These two halves and the new edge form three pieces of the new route to be formed.
All the eight possible ways to combine the three pieces are checked.
The position and the combination that gives the least deadheading cost is selected.
Note that the cost of servicing the edges will be the same irrespective of which combination is selected.
Computing the eight combinations is done in constant time using the cost function.
As there are $l-1$ positions where an edge can be inserted, the computational complexity is $\mc O(l)$, where $l$ is the number of service edges in the route.

\section{Application Scenarios for CAOP}
\label{sc:experiment}
We demonstrate the correlated arc orienteering problem (CAOP) on two applications.

\subsection{Gas Leak Estimation}
We solve the CAOP for the problem of planning routes for estimating gas leak rates~\cite{AlbertsonHFXXFAMBT16,KalvikA22}.
Consider an oil field with multiple oil wells that may leak methane gas with different leak rates.
A ground robot is to efficiently traverse a road network in the oil field and gather methane gas concentration data to estimate gas leak rates; this can be viewed as an informative path planning problem.
Albertson et al.~\cite{AlbertsonHFXXFAMBT16} used a gas dispersion model to compute the posterior distribution of the leak rates given the methane gas concentration data.
They used the expected entropy reduction (EER) information metric to find maximally informative paths.
For our simulation, each oil well is assigned a random leak rate, assumed to be constant for the duration of the experiment. Following Kalvik and Akella~\cite{KalvikA22}, we assume Gaussian priors for the leak rates, which they show lead to an analytical EER and posterior for the leak rates.

For a single oil well, the EER of an edge is a measure of the mutual information provided by gas concentration measurements along the edge about the leak rate of the well.
For a set of wells, we use the sum of their EERs as the reward for the edge.
We wish to find routes that maximize their total EER. %

\begin{figure}[htbp]
	\begin{center}
	\subfloat[AOP: optimal MILP route]{%
	\includegraphics[width=0.33\textwidth, trim={0.8cm 0.5cm 1.5cm 0.5cm},clip]{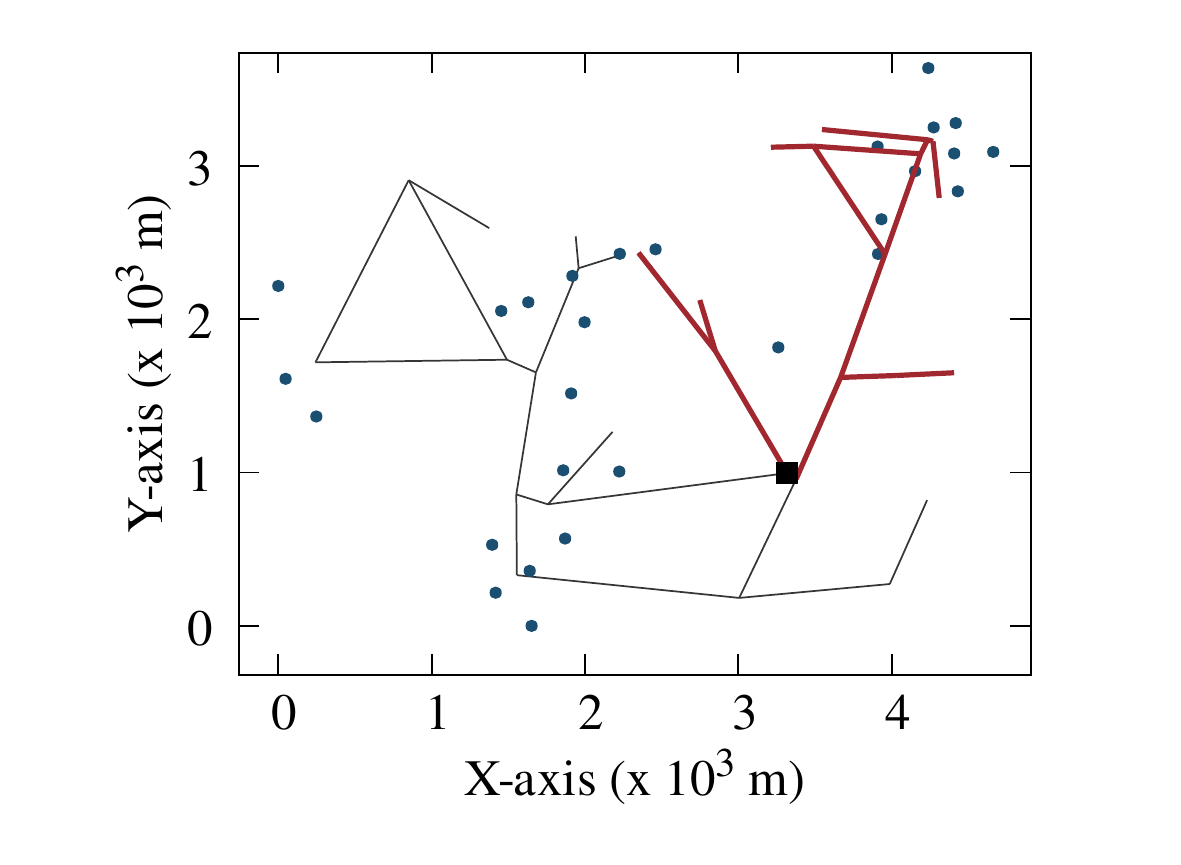}}
	\hfill
	\subfloat[CAOP: optimal MIQP route]{%
	\includegraphics[width=0.33\textwidth, trim={0.8cm 0.5cm 1.5cm 0.5cm},clip]{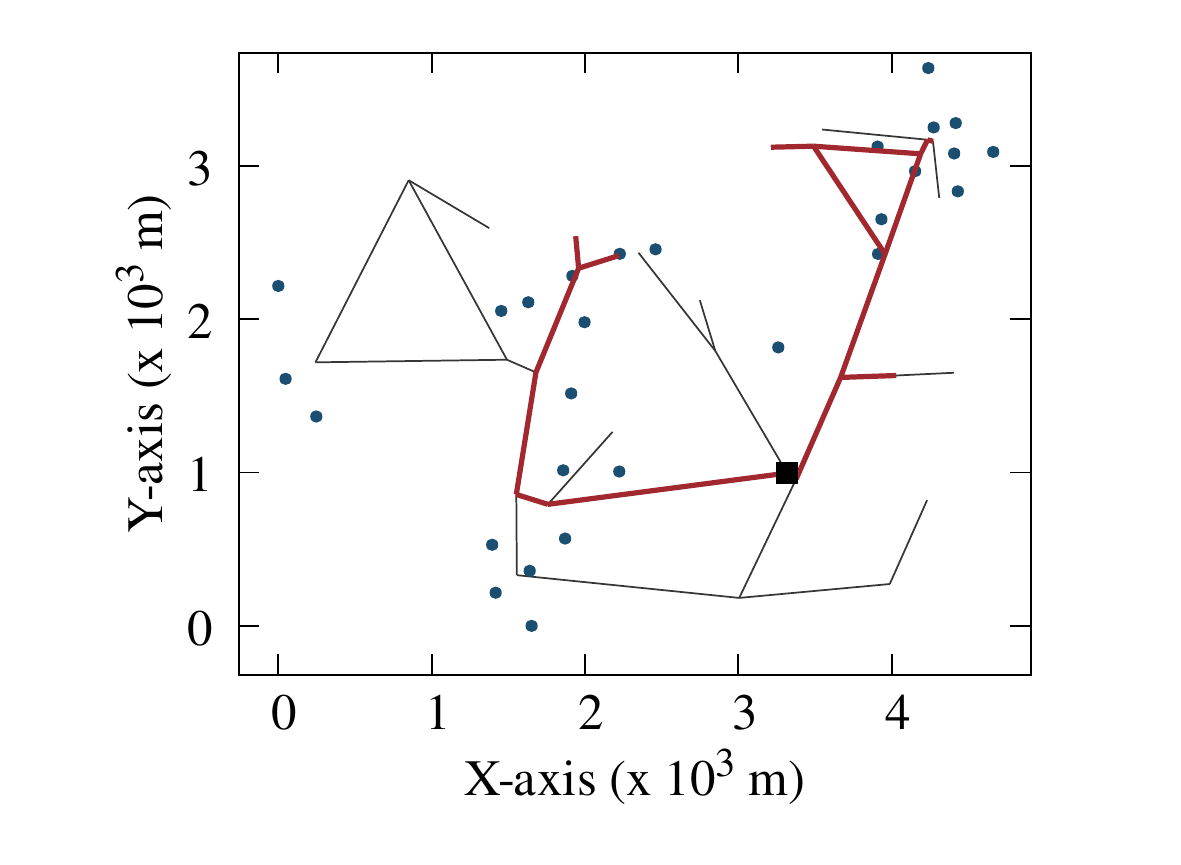}}
	\hfill
	\subfloat[CAOP: greedy route]{%
	\includegraphics[width=0.33\textwidth, trim={0.8cm 0.5cm 1.5cm 0.5cm},clip]{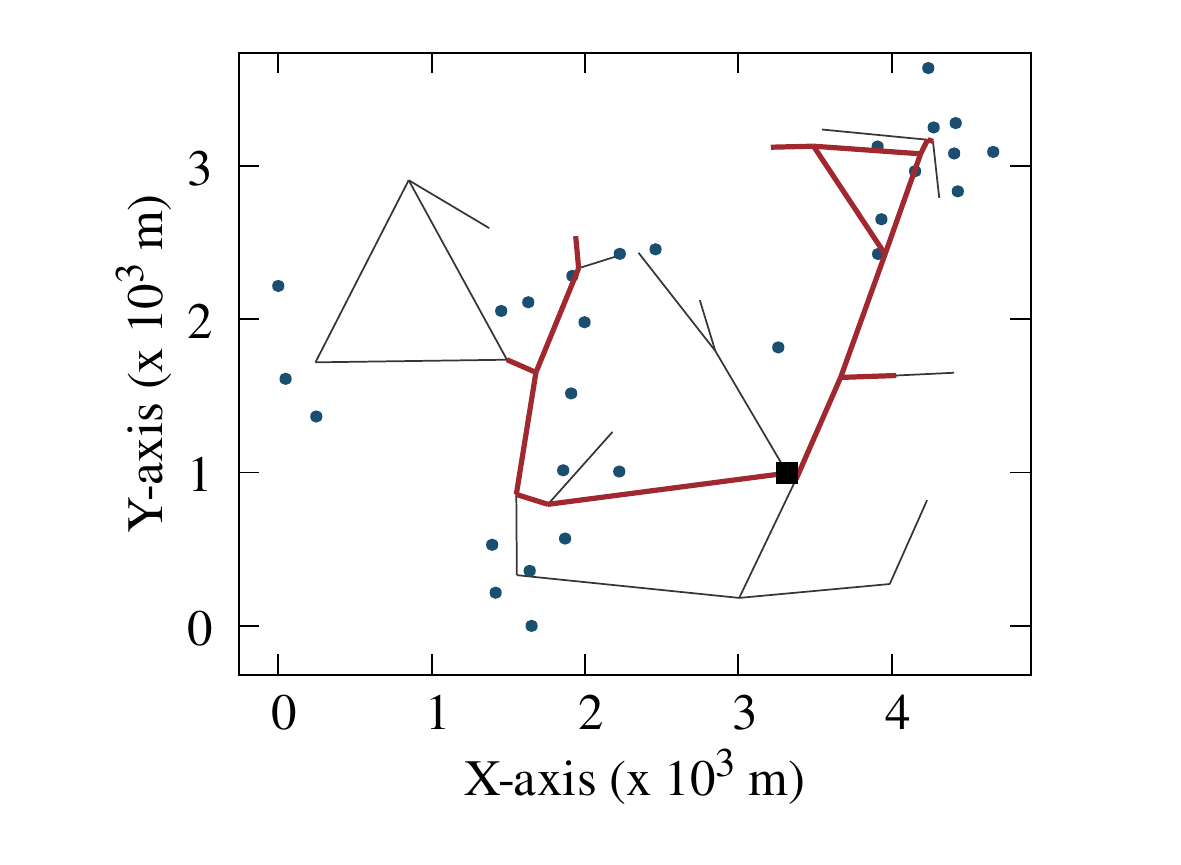}}
	\hfill
	\caption{Routes for methane gas leak detection: The selected routes are drawn bold red, and the underlying road network is shown in gray.
		The blue dots show the locations of the oil wells, and the black square is the depot location.
		The CAOP routes select edges that are less correlated, resulting in a larger total reward.
	The optimal MIQP and the greedy routes are comparable.}
	\label{fig:methane}
\end{center}
\end{figure}

We construct the correlation function~$w$ based on the expected squared distances between edges.
Specifically, the square root of the expected squared distance between points on a pair of edges is computed, denoted by~$d$.
To ensure the correlation matrix values are clamped to a maximum of~1, the distance values for pairs of edges that are very close to each other, i.e., with $d < 1$, are replaced by~1.
The inverse of the distance~$d$ is then computed for each pair of edges.
The largest of these inverses is then used to normalize the entire correlation matrix.
This correlation matrix captures the idea that edges close to each other are highly correlated as they are likely to contain similar information, and the correlation decreases as the distance between two edges increases.

\begin{figure}[tpb]
	\centering
	\begin{tikzpicture}
	\begin{axis}[
		enlargelimits=true,
		xlabel = {Number of edges},
		ylabel = {Reward difference \%},
		xmajorgrids=true,
		grid style=dashed,
		xtick={0,20,40,60,80,100},
		height=4.8cm,
		width=8cm,
		ymin=0,
		ylabel near ticks
		]
		\addplot[
		only marks,
		color=mBlue,
		mark=*,
		mark size=1.0pt]
		table[x=m,y=d, col sep=comma]
		{./graphics/cost_compare.csv};
	\end{axis}
\end{tikzpicture}
	\caption{Comparison of the solutions generated using the greedy heuristic algorithm with the solutions obtained from the MIQP formulation for an application of estimation of gas leaks from oil wells.
	The dataset consists of 100 road networks from the Permian Basin, Texas, USA.}
	\label{fig:cost}
\end{figure}
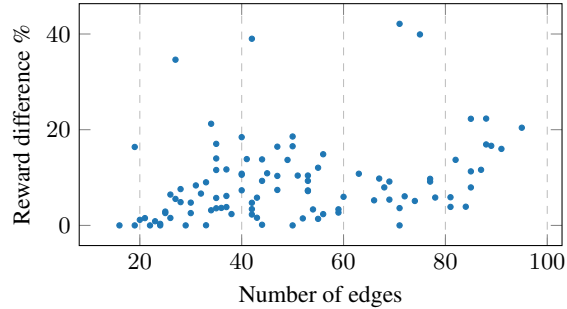

We compared the solutions generated using the greedy heuristic algorithm with the MIQP optimal solutions for a dataset~\cite{KalvikA22} of 100 road networks from the Permian Basin, Texas, USA.
The road networks are composed of up to 100 segments each.
\fgref{fig:methane} shows routes for a road network comprising 33 segments.
The heuristic solutions were within 0\% and 43\% of optimal, and on average were within 9\% of the optimal solutions (see \fgref{fig:cost}).
The heuristic algorithm was executed on a standard laptop with an Intel i7-1195G7 processor, and it computed solutions within 8\,ms for each instance.

\subsection{Coverage of Road Networks}
\begin{figure}[htbp]
	\centering
	\hfill
	\subfloat[Delhi, India]{%
	\includegraphics[width=0.49\textwidth, trim={0.3cm 0.5cm 0.5cm 0.5cm},clip]{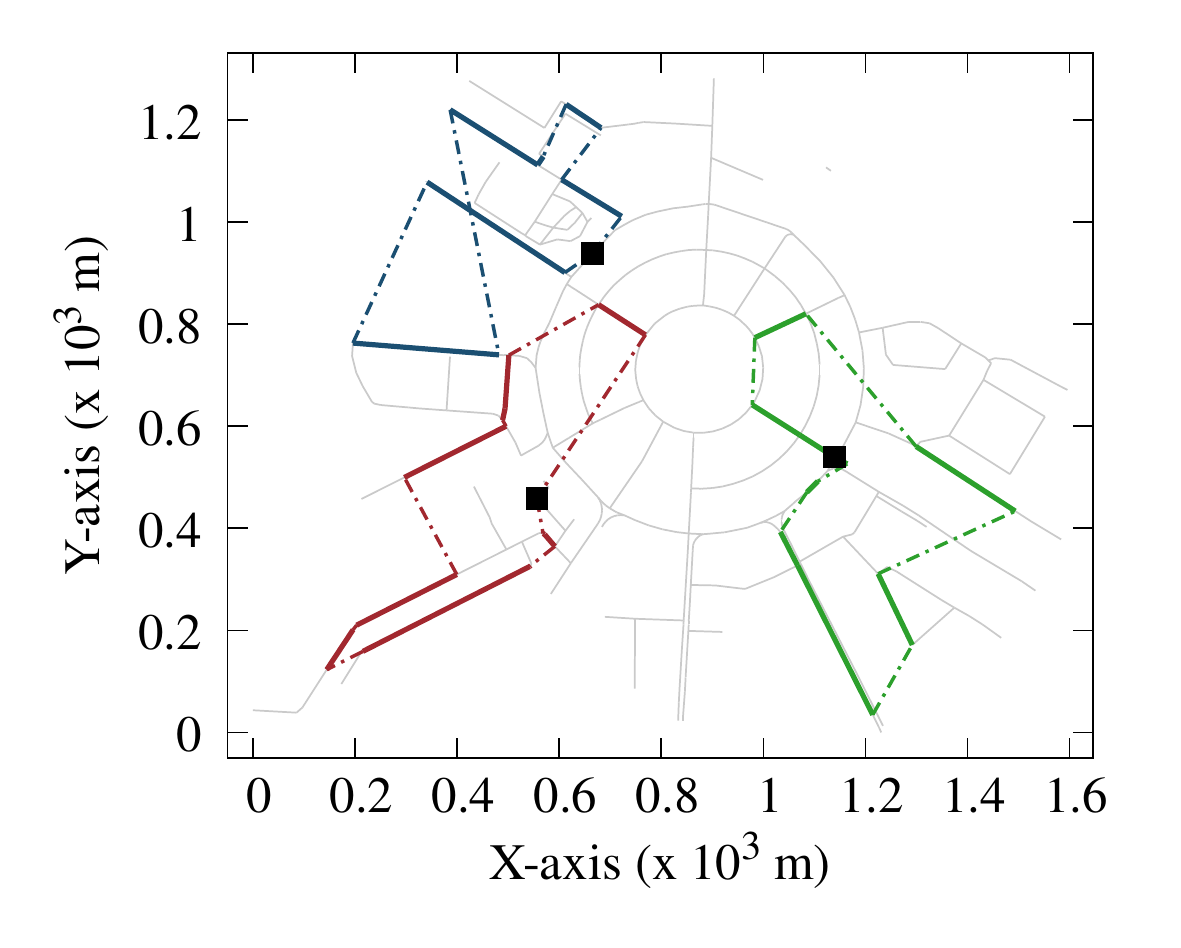}}
	\hfill
	\subfloat[Paris, France]{%
	\includegraphics[width=0.44\textwidth, trim={0.3cm 0.5cm 0.5cm 0.5cm},clip]{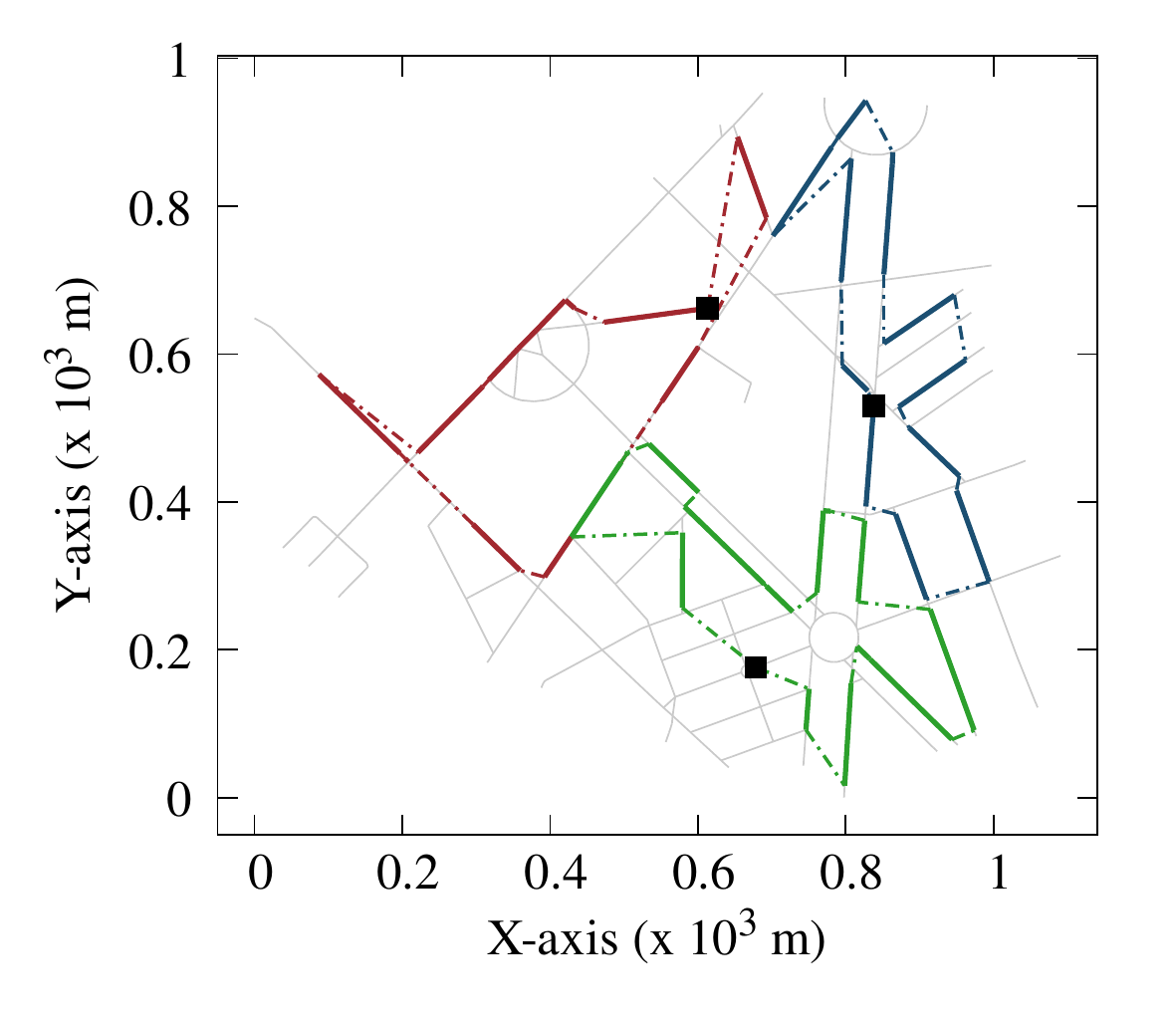}}
	\hfill
	\caption{Routes for three UAVs, for coverage of urban road networks.
		The computed route for each robot is bold and colored, and the underlying road network is shown in light gray.
		The deadheading traversal is shown with dashed lines.
		The CAOP exploits correlations, defined by a field-of-view of 80\,m, to generate routes that cover edges further away from each other.
	\label{fig:delhi}}
\end{figure}

We now illustrate the use of the CAOP for coverage of road networks with a team of uncrewed aerial vehicles (UAVs).
The task is to maximize the coverage of road segments while respecting the battery-constrained flight time of the UAVs.
Since UAVs flying at high altitudes have a large sensor field-of-view, portions of nearby road segments can be observed while traversing a road segment.
The fraction of the road segment observed is used as the correlation function, while the length of the road segment models the associated reward.
The UAVs are modeled to fly at the speed of 3\,m/s when servicing an edge.
As UAVs can fly from one vertex to another, an edge that can only be used for direct deadheading is added for each pair of vertices.
Since the sensors are not used during deadheading, UAVs can fly at higher speeds without concerns of motion blur and sampling rate associated with taking images.
The deadheading speed is set to 5\,m/s, and the flight time for each UAV is set to 600\,s.
\fgref{fig:delhi} illustrates example solutions with three UAVs for two urban road networks.
The Delhi road network comprises 467 vertices, 491 road segments, and 108,\,320 direct edges for deadheading.
These numbers for the Paris road network are 452 vertices, 494 edges, and 101,\,432 directed edges for deadheading.
The depot locations were computed by clustering the edges using $k$-medoids clustering.
The routes were computed by the greedy constructive algorithm, along with clustering for depots, in less than 1.3\,s for each road network.

\section{Conclusion}
\label{sc:conclusion}
This paper introduced the correlated arc orienteering problem (CAOP)---the task of finding routes for multiple resource-constrained robots such that the total reward gathered while traversing environment features is maximized by exploiting correlations between features.
The problem can handle linear and point environment features, multiple robots with individual depots and capacities, and two distinct modes of travel---servicing and deadheading.
The CAOP is suitable for applications in coverage of environments, where rewards are associated with the value of the gathered feature data, and the sensor field-of-view determines the correlations.
Similarly, in informative path planning, rewards are information metrics such as mutual information, and correlations arise from the observed underlying phenomenon.
The CAOP generalizes the correlated orienteering problem (COP) and the team orienteering arc routing problem (TOARP), as the COP considers rewards only on the vertices and the TOARP does not consider the correlations of features.
Furthermore, we relax the constraint in COP that the sum of the weights of neighboring edges is no greater than one.
We developed an MIQP formulation for the CAOP to obtain optimal solutions.
Since the problem is NP-hard, we designed a greedy constructive algorithm.
The greedy and constructive nature of the algorithm makes the algorithm fast.
The algorithm can also be used to obtain a good initial solution for the MIQP.
We demonstrated the MIQP and the algorithm on two applications: methane gas leak detection and environment coverage.

The algorithm is versatile, enabling extensions to several problem variations.
The routing algorithm takes a depot location as input; hence, each potential depot location can be checked to find the depot assignment with the lowest cost in each algorithm iteration.
The routing algorithm can also incorporate nonholonomic constraints without affecting the computational complexity.
In the algorithm, we recomputed the utility for only the neighbors and the co-neighbors of the added edge.
The algorithm can be modified to recompute the utility for all the edges by modifying the correlation matrix to enable higher-order correlations.
Since the CAOP generalizes the COP and the TOARP, with minor modifications, the algorithm can be applied to their corresponding variants.

\section*{Acknowledgments}
\vspace{-0.1in}
{\small We thank Kalvik Jakkala for providing methane leak data and for helpful discussions. This work was supported in part by NSF Award IIP-1919233.}

\bibliographystyle{splncs03}

\end{document}